\documentclass[english]{article}
\usepackage[T1]{fontenc}
\usepackage[latin9]{inputenc}
\usepackage{url}
\usepackage{amsmath}
\usepackage{amssymb}
\usepackage[numbers]{natbib}

\makeatletter

\newcommand{\noun}[1]{\textsc{#1}}

\usepackage[preprint]{nips_2018}

\usepackage{tikz}
\usetikzlibrary{arrows,shapes,positioning,shadows,trees}

\tikzset{
  basic/.style  = {draw, text width=2cm, drop shadow, rectangle},
  root/.style   = {basic, rounded corners=2pt, thin, align=center, fill=gray!40},
  level 2/.style = {basic, rounded corners=6pt, thin,align=center, fill=gray!20, text width=10em},
  level 3/.style = {basic, thin, align=left, fill=white, text width=6.5em}
}

\makeatother

\usepackage{babel}
\begin{document}

\title{Prototype-based Neural Network Layers: Incorporating Vector Quantization}

\author{Sascha~Saralajew\\   Dr. Ing. h.c. F. Porsche AG\\   Weissach, Germany \\   \texttt{sascha.saralajew@porsche.de} \\   \And   Lars Holdijk  \\ Dr. Ing. h.c. F. Porsche AG\\   Weissach, Germany \\   \texttt{lars.holdijk@porsche.de}\\ \And   Maike Rees  \\ Dr. Ing. h.c. F. Porsche AG\\   Weissach, Germany \\   \texttt{maike.rees@porsche.de}\\  \And   Thomas~Villmann\\   University of Applied Sciences Mittweida\\    Mittweida, Germany \\   \texttt{thomas.villmann@hs-mittweida.de}}
\maketitle
\begin{abstract}
Neural networks currently dominate the machine learning community
and they do so for good reasons. Their accuracy on complex tasks such
as image classification is unrivaled at the moment and with recent
improvements they are reasonably easy to train. Nevertheless, neural
networks are lacking robustness and interpretability. Prototype-based
vector quantization methods on the other hand are known for being
robust and interpretable. For this reason, we propose techniques and
strategies to merge both approaches. This contribution will particularly
highlight the similarities between them and outline how to construct
a prototype-based classification layer for multilayer networks. Additionally,
we provide an alternative, prototype-based, approach to the classical
convolution operation. Numerical results are not part of this report,
instead the focus lays on establishing a strong theoretical framework.
By publishing our framework and the respective theoretical considerations
and justifications before finalizing our numerical experiments we
hope to jump-start the incorporation of prototype-based learning in
neural networks and vice versa. 
\end{abstract}

\section{Introduction}

Neural networks (NNs) are currently the state of the art for a wide
range of complex machine learning problems like image classification
or regression tasks. Empowered by hardware support as well as developments
in normalization techniques \cite{Ioffe2015}, activation functions
\cite{GlorotBengio:DeepSparseRectifierNetworks:ProcAISTATS2011,RamachandranEtAl:SwishSelfGatedActivationFunction:arXiv2018}
and new architectures \cite{Redmon2015,Krizhevsky2012,He2015}, it
is hard to ignore NNs. Another reason for their success is that they
are often trained in an end-to-end manner. This means that there are
no intermediate outputs but the network approximates a function that
maps from the input directly to the output decision. The end-to-end
approach however also comes with some disadvantages, such as making
the network hard to interpret, especially in real time. This has led
to a lot of hesitation in implementing NNs in safety critical systems,
such as autonomous cars. This hesitation is further increased by their
lack of robustness. A trained model can be vulnerable to adversarial
attacks \cite{Szegedy2013}. In an adversarial attack, the network
is presented an input with a perturbation that is imperceptible for
the human eye, causing the model to misclassify the input. Despite
the recent efforts of the community and the progress made in the understanding
of these pitfalls, even resulting in provable robustness guarantees
\cite{HeinAndriushchenko:FormalGuaranteesOfClassifierRobustnessAgainstAdversarialAttacks:NIPS2017},
to this day even the best methods are not as robust as humans in classifying
handwritten digits from the MNIST dataset \cite{SchottEtAl:AdversariallyRobustNNforMNIST:2018}.\footnote{Surprisingly, if we follow the results in \cite{SchottEtAl:AdversariallyRobustNNforMNIST:2018},
one of the most robust methods on MNIST is a simple k-nearest-neighbor
classifier, which is a predecessor of Learning Vector Quantization.}

Prototype-based vector quantization methods (PBs) are generally known
for being interpretable, robust, sparse in the sense of memory complexity
and fast to train and to evaluate. PBs rely on the concept of a distance/\,dissimilarity\footnote{We acknowledge that there are differences between mathematical distances
and dissimilarities but will use the words interchangeably for simplicity
in this contribution. For a detailed consideration we refer to \cite{PekalskaDuinBook2006,Villmann2017b}.} measure for classification, giving the approach an inherent geometrically
interpretable robustness. Despite PBs being proposed over 20 years
ago, these algorithms, and particularly their recent improvements
and extensions, are widely unknown or ignored by the current machine
learning community. An exception are the quite simple and frequently
impracticable k-means or k-nearest-neighbor algorithms. Drawbacks
of PBs are that they frequently cannot compete with the accuracy of
modern (deep) neural network architectures. If the task (data) is
complex, like in image classification, the direct input of the image
to the classification is not suitable and requires image features
to be computed a-priori.

Our contribution in this technical report is to bridge the gap between
NNs and PBs. Right now, both methods are treated as completely different,
or even worse as competing or not compatible with each other. This
technical report will show that this is not the case and that the
methods are more related to each other than one might expect at the
first glance. We will not present numerical results, but rather present
a conceptual framework on how the methods relate to each other. Hopefully,
this is a starting point to develop new architectures with benefits
from the two ``worlds''.

As we aim to bring together both research communities we begin this
paper with an overview of the theoretical backgrounds of NNs and PBs.
In chapter four we show a different interpretation of fully-connected
layers by reformulating the underlying operations. Chapter five takes
a similar approach but applies it to convolutional layers. Chapter
six will introduce a prototype-based classification layer and a prototype
inspired version of the convolution operation, often applied in NNs.
Both methods rely on the different interpretations explored in chapter
four and five. We discuss their benefits in chapter seven together
with empirically gained insights where possible. Tips and tricks,
discovered during our ongoing experiments are described in chapter
eight. Chapter nine presents related approaches and chapter ten concludes
and gives an overview of planned future work.

\section{Introduction to prototype-based methods}

\begin{figure}
\begin{centering}
\begin{tikzpicture}[   
    level 1/.style={sibling distance=75mm},   
    level 2/.append style={sibling distance=45mm},   
    edge from parent/.style={->,draw, black},   >=latex] 

% root of the the initial tree, level 1 
\node[root] {\textbf{Vector quantization and clustering}}
% The first level, as children of the initial tree   
    child {node[level 2] (ch1) {Non prototype-based}        
    }   
    child {node[level 2] (ch2) {Prototype-based (PB)}     
        child {node[level 2] (c4) {Unsupervised (VQ)}}     
        child {node[level 2] (c5) {Supervised (LVQ)}}        
    }; 
% The second level, relatively positioned nodes 
\begin{scope}[every node/.style={level 3}] 
\node [below = of  ch1, xshift=15pt, yshift=10pt] (c11) {Decision Trees};
\node [below = of  c11, yshift=20pt] (c12) {Affinity Propagation};
\node [below = of  c12, yshift=20pt] (c13) {Hierarchical Clustering};
\node [below = of  c13, yshift=20pt] (c14) {DBSCAN};
\node [below = of  c14, yshift=20pt] (c15) {...};

\node [below = of  c4, xshift=15pt, yshift=10pt] (c41) {k-means};  
\node [below = of  c41, yshift=20pt] (c42) {Fuzzy k-means};
\node [below = of  c42, yshift=20pt] (c43) {Neural gas}; 
\node [below = of  c43, yshift=20pt] (c44) {Self-Organizing Maps};
\node [below = of  c44, yshift=20pt] (c45) {...}; 

\node [below = of  c5, xshift=15pt, yshift=10pt] (c51) {k-nearest-neighbors};  
\node [below = of  c51, yshift=20pt] (c52) {LVQ2.1};  
\node [below = of  c52, yshift=20pt] (c53) {GLVQ}; 
\node [below = of  c53, yshift=20pt] (c54) {RSLVQ};
\node [below = of  c54, yshift=20pt] (c55) {GMLVQ}; 
\node [below = of  c55, yshift=20pt] (c56) {...};

\end{scope} 

% lines from each level 1 node to every one of its "children" 
\foreach \value in {1,...,5}
    \draw[->] (ch1.195) |- (c1\value.west); 

\foreach \value in {1,...,5}   
    \draw[->] (c4.195) |- (c4\value.west);

\foreach \value in {1,...,6}
    \draw[->] (c5.195) |- (c5\value.west); 
\end{tikzpicture}
\par\end{centering}
\caption{An overview of the general vector quantization/\,clustering methods
landscape, explaining how different approaches relate to each other.
It is important to note here that the unsupervised prototype-based
vector quantization classifier referred to in this contribution is
in fact only a part of the encompassing vector quantization methodology.
For simplicity we have decided to use the name vector quantization
and its abbreviation VQ to refer to the unsupervised prototype-based
variant, as is done in most related work. \label{fig:An-overview-of}}
\end{figure}
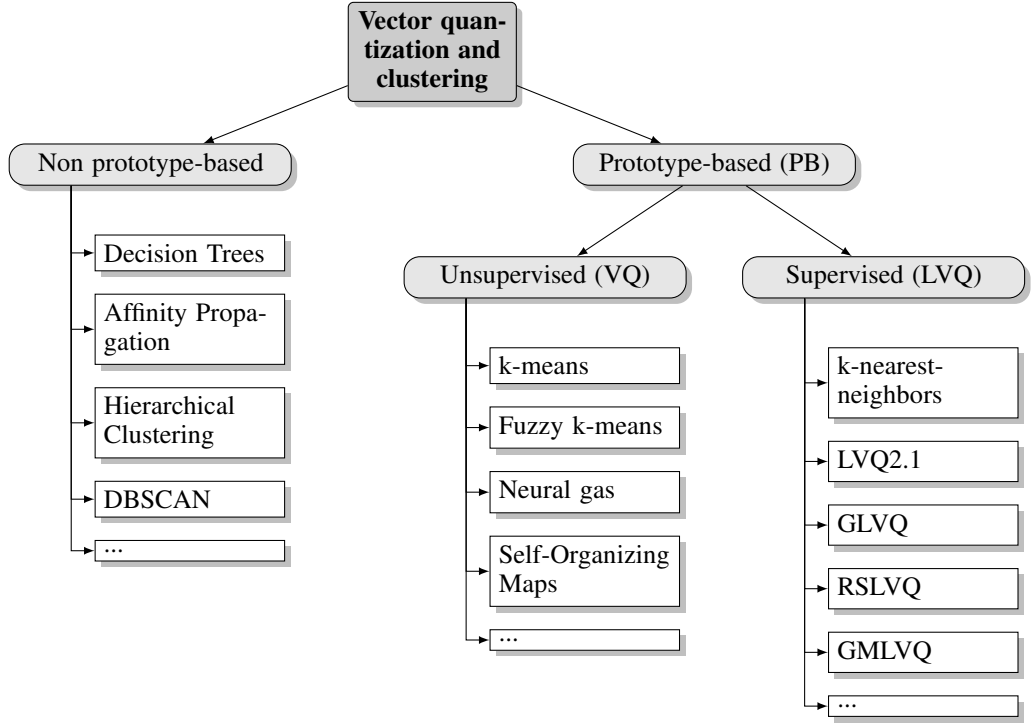
Vector quantization is one of the most successful approaches for data
clustering and representation as well as compression \cite{Duda73a,Bishop2006a}.
One of the pioneers in vector quantization,\noun{ R.\,M. Gray} stated
\cite{Gray:VectorQuantization:IEEEASSPMagazine:1984}: \emph{``A
vector quantizer is a system for mapping a sequence of continuous
or discrete vectors into a digital sequence suitable for communication
over or storage in a digital channel. The goal of such a system is
data compression: to reduce the bit rate so as to minimize communication
channel capacity or digital storage memory requirements while maintaining
the necessary fidelity of the data.''}

Most of the methods use the concept of prototypes and are therefore
called PBs \cite{Villmann2016e}. These methods are based on a set
$W=\left\{ \mathbf{w}_{1},\mathbf{w}_{2},...,\mathbf{w}_{N_{W}}\right\} $
of prototypes $\mathbf{w}_{k}$, usually in the $\mathbb{R}^{n}$,
and a suitable dissimilarity or distance measure $d$ \cite{Villmann2017b}.
The\emph{ prototype response vector} is defined as
\begin{equation}
\mathbf{d}\left(\mathbf{x}\right)=\left(d\left(\mathbf{x},\mathbf{w}_{1}\right),\ldots,d\left(\mathbf{x},\mathbf{w}_{N_{W}}\right)\right)^{\textrm{T}}\label{eq:prototype response vector}
\end{equation}
with the respective dissimilarity measure. Frequently, this is defined
as the squared Euclidean metric
\begin{equation}
d_{E}^{2}\left(\mathbf{x},\mathbf{w}\right)=\left(\mathbf{x}-\mathbf{w}\right)^{\textrm{T}}\left(\mathbf{x}-\mathbf{w}\right).\label{eq:Euclidean-distance}
\end{equation}
The vector quantization mapping $\mathbf{x}\mapsto\kappa\left(\mathbf{x}\right)$
takes place as a winner-takes-all (WTA) rule according to
\begin{equation}
\kappa\left(\mathbf{x}\right)=\arg\min\left\{ d\left(\mathbf{x},\mathbf{w}_{k}\right)|k=1,2,...N_{w}\right\} .\label{eq:wta}
\end{equation}
realizing the nearest prototype principle with respect to the given
dissimilarity. The value $\kappa\left(\mathbf{x}\right)$ is called
the index of the best matching (winner) prototype. According to the
assignments $\kappa\left(\mathbf{x}\right)$, a tessellation of the
data space is induced by the Voronoï cells 
\begin{equation}
V_{k}=\left\{ \mathbf{x}\in\mathbb{R}^{n}|k=\kappa\left(\mathbf{x}\right)\right\} \label{eq:Voronoi cell}
\end{equation}
assigned to each prototype $\mathbf{w}_{k}$. It is frequently denoted
as the receptive field of neural vector quantizers \cite{Voronoi1908a}. 

For unsupervised prototype-based vector quantization (VQ), each prototype
is assumed to serve as a cluster center and hence, $\kappa\left(\mathbf{x}\right)$
of the WTA-rule (\ref{eq:wta}) delivers the cluster index. A model
is trained either by a heuristically motivated update of the prototypes
or by the optimization of a cost function \cite{Zador82a,yair92a,Linde1980a}.
After the training, the prototypes should be distributed in the data
space such that the training dataset $\mathcal{X}\subseteq\mathbb{R}^{n}$
is well covered. Probably the most prominent example of VQ algorithms
is k-means \cite{Linde1980a,LloydKMeans1982}. In the past, several
extensions and improvements of VQ methods were invented. For example
methods with improved convergence behavior \cite{kohonen95a,ArthurVassilvitskii:Kmeansplusplus:ProcACMSIAM:2007},
including neural gas \cite{martinetz93d}, fuzzy variants where the
crisp WTA-rule is replaced by a smoothed decision function \cite{Bezdek81a,Hathaway1989a,PalPalKellerBezdek2005a,Hathaway1994a}
and variants that use a different metric than the Euclidean distance
\cite{Villmann2015a,Villmann2011a}. A systematic overview is given
in \cite{Miyamoto2008_Algorithms_for_Fuzzy_Clustering}. 

The supervised counterpart to VQ is called Learning Vector Quantization
(LVQ). It was initially developed for classification learning by \noun{T.
Kohonen} in 1988 \cite{kohonen88j} and later extended \cite{kohonen95a}.
In LVQ, each prototype $\mathbf{w}_{k}$ is additionally equipped
with a class label $c_{k}\in C=\left\{ 1,2,...,N_{C}\right\} $ and
the prototypes are distributed regarding a training dataset $\mathcal{X}=\left\{ \left(\mathbf{x},c\left(\mathbf{x}\right)\right)|\mathbf{x}\in\mathbb{R}^{n},c\left(\mathbf{x}\right)\in C\right\} $.
The class of a given data point $\mathbf{x}$ is defined as 
\begin{equation}
\mathbf{x}\longmapsto c\left(\mathbf{x}\right)=c_{\kappa\left(\mathbf{x}\right)},\label{eq:LVQ class mapping}
\end{equation}
yielding a crisp classification. Originally, the learning process
of the prototypes was heuristically motivated, later it was refined
to optimize a cost function which approximates the overall classification
accuracy \cite{sato96a,Villmann2014e}. Probabilistic variants of
LVQ are the Robust Soft LVQ (RSLVQ \cite{Seo2003a}), the soft nearest
prototype classifier \cite{Seo2003b} or the probabilistic LVQ \cite{Villmann2018b}
to name just a few. See Fig.~\ref{fig:An-overview-of} for an incomplete
overview of the vector quantization landscape and its most prominent
variants.

Non-standard metrics are also attractive for PBs \cite{Villmann2015a,Villmann2011b,Villmann2014f,Schneider2009_DistanceLearning}
since their incorporation provides a powerful modification. Generalized
Matrix LVQ (GMLVQ \cite{Schneider2009_MatrixLearning}) supposes an
adjustable dissimilarity measure, which is adapted during the training
process. This adaption takes place in parallel to the prototype learning.
The dissimilarity measure is defined as
\begin{equation}
d_{\boldsymbol{\Omega}}^{2}\left(\mathbf{x},\mathbf{w}\right)=\left(\mathbf{x}-\mathbf{w}\right)^{\textrm{T}}\boldsymbol{\Omega}^{\textrm{T}}\boldsymbol{\Omega}\left(\mathbf{x}-\mathbf{w}\right)\label{eq:omega-diss}
\end{equation}
where the matrix $\boldsymbol{\Omega}\in\mathbb{R}^{m\times n}$ is
learned during the training. We denote this as Omega dissimilarity.
The commonality of all these modifications is that the learning results
in a suitable distribution of the prototypes in the data space.

The distance based perspective in prototype-based methods offers a
clear understanding of the prototype principle and the WTA-rule provides
many interpretation techniques e.\,g. biological excitation in neural
maps or local area winner in Heskes-Self-Organizing-Maps \cite{Willshaw76a,heskes99a,Villmann2018g}.
It should be emphasized that the optimum assignment of a data sample
$\mathbf{x}$ to a prototype $\mathbf{w}_{k}$ is achieved for $d\left(\mathbf{x},\mathbf{w}_{k}\right)=0$,
which can be interpreted as a\emph{ lower bound}. This geometrical
perspective makes the method attractive for outlier detection and
related reject strategies \cite{FischerHammerWersing:EfficientRejectionStrategies:Neurocomputing2015,HerbeiWegkampClassifcationwithRejectCJourStat2006,TradeoffAccuracyRejectChow1970,UrahamaFurukawa:GradienDescentNPCwithReject:PattRec1995,VailayaRejectOptionsBayesianVQClassification2000,YuanWegkampClassificationRejectConvexRiskMinimizationJMLR2010,munoz98a}. 

\section{Introduction to neural networks}

In recent years, NNs and more specifically Convolutional Neural Networks
(CNNs), gained a lot of attention due to their successes in a variety
of difficult classification and regression tasks \cite{Krizhevsky2012,Szegedy2013,He2015}. 

A neural multilayer feedforward network classifier can be decomposed
into a feature extraction network concatenated with a classification
network. The feature extraction network extracts a feature representation
$f\left(\mathbf{x}\right)$ from the input $\mathbf{x}$. This feature
representation is consecutively passed into the network to obtain
the final classification decision. For image classification, the extraction
is usually realized using a stack of convolutional layers. The output
$f\left(\mathbf{x}\right)$ of the extraction network can then be
seen as a feature stack (images of features). A reshaped representation
of the feature stack is then the input layer of the classification
network. Usually, this is defined as a multilayer perceptron network
(fully-connected feedforward network). The ability to train the feature
extractor and classifier at the same time is one of the main strengths
of NNs and was instrumental to the adaption of this machine learning
method. 

In recent years a lot of techniques were developed to improve the
accuracy and ease of training. Examples of these are: dropout \cite{Hinton2012},
batch normalization \cite{Ioffe2015}, improved architecture \cite{Redmon2015,Krizhevsky2012,He2015}
and different activation functions \cite{GlorotBengio:DeepSparseRectifierNetworks:ProcAISTATS2011,RamachandranEtAl:SwishSelfGatedActivationFunction:arXiv2018}.
In addition to this, newly available hardware support by GPU computing
enabled the use of larger and deeper networks. For a more complete
introduction and an overview of the history of NNs we refer to \cite{Schmidhuber2014}.

\section{A different view on fully-connected layers\label{sec:A-different-view}}

In the following, the relation between fully-connected layers (FCLs)
and LVQ networks is explored. This allows to rewrite both of them
using equivalent operations. This strategy is already described for
different levels of abstraction in \cite{deVries:DeepLVQ:ESANN2016,Villmann2017h,OordEtAl:NeuralDiscreteRepresentationLearning:NIPS2017,AgustssonEtAl:SoftHardVQForEnd2EndLearning:NIPS2017,YangEtAl:ClassificationWithConvolutioanlPrototypeLearning:CVPR2018}.

\subsection{Relation between Euclidean distance-based PBs and FCLs\label{subsec:Relation-between-the}}

In the simple form, the output $\mathbf{o}\left(\mathbf{x}\right)$,
regarding a given input $\mathbf{x}$ of a FCL of a feedforward network
with trivial activation $id\left(\mathbf{x}\right)=\mathbf{x}$, can
be written as
\begin{equation}
\mathbf{o}\left(\mathbf{x}\right)=\mathbf{A}\mathbf{x}+\mathbf{b}\label{eq:output fully-connected-perceptron-layer}
\end{equation}
where $\mathbf{A}$ is the weight matrix and $\mathbf{b}$ is the
bias vector. Hence, the components are calculated according to
\begin{equation}
o_{k}\left(\mathbf{x}\right)=\left\langle \mathbf{a}_{k},\mathbf{x}\right\rangle _{E}+b_{k}
\end{equation}
where $\left\langle \mathbf{a}_{k},\mathbf{x}\right\rangle _{E}$
is the Euclidean inner product and $\mathbf{a}_{k}$ denotes the $k$-th
row vector of the matrix $\mathbf{A}$.

Suppose that there is exactly one prototype $\mathbf{w}_{k}$ that
is responsible for each class and that the class $c_{k}$ of this
prototype is $k$, i.\,e. $c_{k}=k$ with $k\in\left\{ 1,\ldots,N_{C}\right\} $
and $N_{W}=N_{C}$.\footnote{The restriction to one prototype per class can be easily relaxed and
a similar formulation can be obtained.} Based on these assumptions, the prototype response vector in (\ref{eq:prototype response vector})
can be written as
\begin{equation}
\mathbf{d}\left(\mathbf{x}\right)=\left(d_{E}^{2}\left(\mathbf{x},\mathbf{w}_{1}\right),d_{E}^{2}\left(\mathbf{x},\mathbf{w}_{2}\right),...,d_{E}^{2}\left(\mathbf{x},\mathbf{w}_{N_{C}}\right)\right)^{\textrm{T}}\label{eq:native-GLVQ-formulation of prototype response}
\end{equation}
for the squared Euclidean distance measure.

However, this formulation (\ref{eq:native-GLVQ-formulation of prototype response})
of the prototype response shows a\emph{ computational disadvantage}:
for parallel computation of the response vector $\mathbf{d}\left(\mathbf{x}\right)$,
multiple copies of the input $\mathbf{x}$ are needed, in order that
each component $d_{k}\left(\mathbf{x}\right)=d_{E}^{2}\left(\mathbf{x},\mathbf{w}_{k}\right)$
can be calculated independently. Hence, the respective storage requirements
grow with the number of classes/\,prototypes. This bottleneck of
the naive approach limits the scalability of LVQ to large networks
or to networks with a huge number of classes.

By considering the identity
\begin{equation}
d_{E}^{2}\left(\mathbf{x},\mathbf{w}_{k}\right)=\left\langle \mathbf{x},\mathbf{x}\right\rangle _{E}-2\left\langle \mathbf{x},\mathbf{w}_{k}\right\rangle _{E}+\left\langle \mathbf{w}_{k},\mathbf{w}_{k}\right\rangle _{E}\label{eq:identity-of-dE-to-dot-product}
\end{equation}
for the Euclidean distance, these limitations can be overcome. For
this purpose, all prototypes are collected into a matrix $\mathbf{W}\in\mathbb{R}^{N_{C}\times n}$
in order that
\[
\mathbf{W}=\left(\mathbf{w}_{1}|\mathbf{w}_{2}|...|\mathbf{w}_{N_{C}}\right)^{\textrm{T}}
\]
is valid. Now Eq.~(\ref{eq:native-GLVQ-formulation of prototype response})
is equivalent to
\begin{equation}
\mathbf{d}\left(\mathbf{x}\right)=-2\mathbf{W}\mathbf{x}+\mathbf{b}\left(\mathbf{x},\mathbf{W}\right)\label{eq:efficient-GLVQ-formulation prototype response}
\end{equation}
with
\begin{equation}
\mathbf{b}\left(\mathbf{x},\mathbf{W}\right)=\left(\left\Vert \mathbf{x}\right\Vert _{2}^{2}+\left\Vert \mathbf{w}_{1}\right\Vert _{2}^{2},\left\Vert \mathbf{\mathbf{x}}\right\Vert _{2}^{2}+\left\Vert \mathbf{w}_{2}\right\Vert _{2}^{2},...,\left\Vert \mathbf{x}\right\Vert _{2}^{2}+\left\Vert \mathbf{w}_{N_{C}}\right\Vert _{2}^{2}\right)^{\textrm{T}}\label{eq:bias vector for effective prototype response}
\end{equation}
and thus the parallel calculation of $\mathbf{b}\left(\mathbf{x},\mathbf{W}\right)$
requires only copies of the single value $\left\Vert \mathbf{x}\right\Vert _{2}^{2}$
instead of the full vectors $\mathbf{x}$.

It can be observed that the computation of the response vector $\mathbf{d}\left(\mathbf{x}\right)$
according to (\ref{eq:efficient-GLVQ-formulation prototype response})
is structurally equivalent to the output (\ref{eq:output fully-connected-perceptron-layer})
of a FCL with trivial activation $id\left(\mathbf{x}\right)=\mathbf{x}$:
Both methods apply a linear transformation to the input vector and
add a bias term. The difference is that in a FCL (\ref{eq:output fully-connected-perceptron-layer})
the bias term $\mathbf{b}$ is a trainable parameter, whereas in LVQ
the bias term $\mathbf{b}\left(\mathbf{x},\mathbf{W}\right)$ depends
dynamically on the prototypes and the current input vector.

\subsection{Relation between the Omega dissimilarity and two FCLs \label{subsec:Omega distance and FCL} }

A multilayer perceptron that combines multiple FCLs can be seen as
a repeated application of Eq.~(\ref{eq:output fully-connected-perceptron-layer}),
using a generally nonlinear activation function $\phi$ (e.\,g. the
Rectified Linear Unit - ReLU \cite{GlorotBengio:DeepSparseRectifierNetworks:ProcAISTATS2011}
or sigmoid).

Consider a network with two FCLs and trivial activation in the second
layer. The output can be defined as
\begin{equation}
\mathbf{o}\left(\mathbf{x}\right)=\mathbf{A}\boldsymbol{\phi}\left(\mathbf{A}_{1}\mathbf{x}+\mathbf{b}_{1}\right)+\mathbf{b}\label{eq:multilayer-full-connected}
\end{equation}
where $\mathbf{A}_{1}$ and $\mathbf{b}_{1}$ are the parameters of
the first layer. This shows that the first layer applies an affine
transformation of the input vector $\mathbf{x},$ followed by a transformation
according to the activation $\boldsymbol{\phi}\left(\mathbf{x}\right)=\left(\phi\left(x_{1}\right),\ldots,\phi\left(x_{n}\right)\right)^{\textrm{T}}$.
Afterwards, the vector $\boldsymbol{\phi}\left(\mathbf{x}\right)$
serves as the input to the second layer, which delivers the output
by following the rule (\ref{eq:output fully-connected-perceptron-layer}).
Hence, the first layer can be interpreted as nonlinear input transformation
of $\mathbf{x}$ due to the nonlinear activation function.

For comparison, consider a variation of the Omega dissimilarity $d_{\mathbf{\Omega}}$
of GMLVQ from (\ref{eq:omega-diss}), i.\,e. the measure
\begin{equation}
\delta_{\boldsymbol{\Omega}}^{2}\left(\mathbf{x},\mathbf{w}\right)=\left(\boldsymbol{\Omega}\mathbf{x}-\mathbf{w}\right)^{\textrm{T}}\left(\boldsymbol{\Omega}\mathbf{x}-\mathbf{w}\right)\label{eq:reformualted-Omega-diss}
\end{equation}
is investigated. The data vector $\mathbf{x}$ is first linearly projected
by the projection matrix $\boldsymbol{\Omega}$ into the projection
space and consequently the prototypes are defined within this space.
The initial projection $\boldsymbol{\Omega}\mathbf{x}$ can be interpreted
as a FCL without bias and with trivial activation. With the application
of Eq.~(\ref{eq:efficient-GLVQ-formulation prototype response})
still being valid, the application of $\delta_{\boldsymbol{\Omega}}$
can be understood as a two FCL network with the constraints described
above. Furthermore, this perspective can be generalized to the measure
\begin{equation}
\mu_{\boldsymbol{\Omega}}^{2}\left(\mathbf{x},\mathbf{w}\right)=\left(\boldsymbol{\phi}\left(\boldsymbol{\Omega}\mathbf{x}-\mathbf{b}\right)-\mathbf{w}\right)^{\textrm{T}}\left(\boldsymbol{\phi}\left(\boldsymbol{\Omega}\mathbf{x}-\mathbf{b}\right)-\mathbf{w}\right)\label{eq:nonlinear projection distance}
\end{equation}
as suggested in \cite{Villmann2019a}. In this way a complete correspondence
is obtained between the Omega dissimilarity and a multiplayer perceptron
using two FCLs.

\section{A different view on convolutional layers\label{sec:A-different-view-on-CNNs}}

In the following, we show how a matrix multiplication can be expressed
in terms of convolution operations. In CNNs the convolutional layer
is not defined as a mathematical convolution, instead it is a mathematical
cross-correlation also denoted as sliding dot product (inner product)
\cite{dumoulin2016guide}. Nevertheless, we will refer to it as convolution
for simplicity.

In this chapter we suppose that $\mathbf{x}\in\mathbb{R}^{w\times h\times c}$
is a feature stack of size $w\times h\times c$ and $\mathbf{k}\in\mathbb{R}^{w_{f}\times h_{f}\times c}$
a convolutional kernel. Suppose that $\mathbf{x}_{ij}\in\mathbb{R}^{w_{f}\times h_{f}\times c}$
is the window at the pixel position $i,j$ of $\mathbf{x}$ where
the response with the kernel $\mathbf{k}$ should be evaluated. Then,
the output of the convolution $\mathbf{x}\ast\mathbf{k}$ at pixel
position $i,j$ is given by
\begin{equation}
\left.\mathbf{x}\ast\mathbf{k}\right|_{ij}=\tilde{\mathbf{x}}_{ij}^{\textrm{T}}\tilde{\mathbf{k}}\label{eq:convolution-as-dot-product}
\end{equation}
where $\tilde{\mathbf{x}}_{ij}$ and $\tilde{\mathbf{k}}$ are replicas
of $\mathbf{x}_{ij}$ and $\mathbf{k}$ in vector shape, i.\,e. $\tilde{\mathbf{x}}_{ij}\in\mathbb{R}^{w_{f}\cdot h_{f}\cdot c}$
and $\tilde{\mathbf{k}}\in\mathbb{R}^{w_{f}\cdot h_{f}\cdot c}$. 

Like usual in CNNs, we filter the images with a stack of filters.
Assume there are $N_{f}$ filters of shape $\mathbf{k}$ collected
in a tensor of the shape $w_{f}\times h_{f}\times c\times N_{f}$.
Then, the convolution with all these filters can be seen as a matrix
multiplication. More precisely, it is a linear transformation of the
vector collected over a sliding window from $\mathbb{R}^{w_{f}\cdot h_{f}\cdot c}\rightarrow\mathbb{R}^{N_{f}}$
by
\begin{equation}
\tilde{\mathbf{K}}\tilde{\mathbf{x}}_{ij}\label{eq:convolution-as-matrix-multiplication}
\end{equation}
where $\tilde{\mathbf{K}}=\left(\tilde{\mathbf{k}}_{1}|\tilde{\mathbf{k}}_{2}|...|\tilde{\mathbf{k}}_{N_{f}}\right)^{\textrm{T}}$
is the matrix of the vectorized filters $\tilde{\mathbf{k}}_{l}$.

If the convolution operation is equipped with the additional bias
term $\tilde{b}_{l}$ for each kernel, then this is equivalent to
a linear transformation of the form (\ref{eq:convolution-as-matrix-multiplication})
followed by a vector shift, i.\,e. the resulting operation
\[
\tilde{\mathbf{K}}\tilde{\mathbf{x}}_{ij}+\tilde{\mathbf{b}}
\]
is an affine transformation.

To summarize, the general convolution operation of CNNs is an affine
transformation of each sliding window with the same transformation
(known as shared weights).

\section{Prototype-based neural network layers}

In the previous sections we have shown how NNs and the prototype-based
learning methods LVQ and VQ relate to each other. Using these relations,
we propose two prototype-based layers to be used in NNs in the following
section. 

\subsection{LVQ layers as final classification layers}

As mentioned in Sec.~\ref{sec:A-different-view}, the implementation
of LVQ as final classification network of a NN, split into feature
extraction and classification, is straightforward. During the training
of the NN the prototypes are trained in parallel with the feature
extraction layers, using the output of the feature extraction layer
as input to the prototype-based model. At inference, the distance
between the output of the feature extraction layer and all learned
prototypes is used for classification. This requires the selection
of a differentiable loss function and distance measure. 

Below, two special cases of LVQ classification layers are proposed
and their loss functions are defined.

\paragraph{RSLVQ }

In general, the output vector $\mathbf{o}\text{\ensuremath{\left(\mathbf{x}\right)}}$
of (\ref{eq:output fully-connected-perceptron-layer}) in the last
FCL of a NN is element-wise normalized by the $\textrm{softmax}$
activation (also denoted as Gibbs measure/\,distribution) given 
\[
\hat{\mathbf{p}}\left(\mathbf{x}\right)=\textrm{\textrm{softmax}}\left(\mathbf{o}\text{\ensuremath{\left(\mathbf{x}\right)}}\right)
\]
with $\hat{\mathbf{p}}\left(\mathbf{x}\right)\in\left[0,1\right]^{N_{C}}$
being a probability vector of the estimated class probabilities and
the softmax function defined as
\begin{equation}
\textrm{\textrm{softmax}}\left(z_{k}\right)=\frac{\exp\left(z_{k}\right)}{\sum_{i}\exp\left(z_{i}\right)}\label{eq:softmax}
\end{equation}
for vectors $\mathbf{z}=\left(z_{1},\ldots,z_{n}\right)^{\textrm{T}}$.
The training of the network is usually realized applying the cross
entropy loss for the network class probability $\hat{\mathbf{p}}\left(\mathbf{x}\right)$
and the true class probability $\mathbf{p}\left(\mathbf{x}\right)$. 

Robust Soft Learning Vector Quantization (RSLVQ) was introduced by
\noun{Seo} and \noun{Obermayer} \cite{Seo2003a} as a probabilistic
version of LVQ. RSLVQ defines the prototypes as centers of Gaussian
densities and the whole machine learning model as Gaussian density
mixture model. During the learning, the prototypes are adjusted to
approximate the class probability densities of the data. This is achieved
via a maximum log-likelihood optimization of the class dependent Gaussian
mixture model. For that, the prototype response vector $\mathbf{d}\left(\mathbf{x}\right)$
of distance values from (\ref{eq:efficient-GLVQ-formulation prototype response})
is transformed into a probability vector via
\begin{equation}
\hat{\mathbf{p}}\left(\mathbf{x}\right)=\textrm{\textrm{softmax}}\left(-\mathbf{d}\left(\mathbf{x}\right)\right)\label{eq:neg_softmax}
\end{equation}
in the RSLVQ network, which obeys a Gaussian probability in presence
of the squared Euclidean distance. It turns out that the maximum log-likelihood
loss function of RSLVQ is equivalent to the cross entropy loss between
$\hat{\mathbf{p}}\left(\mathbf{x}\right)$ and $\mathbf{p}\left(\mathbf{x}\right)$
\cite{Villmann2018b}. 

Hence, a RSLVQ classifier can be interpreted as fully-connected $\textrm{\textrm{softmax}}$
classification layer with the particular bias $\mathbf{b}\left(\mathbf{x},\mathbf{W}\right)$
according to (\ref{eq:bias vector for effective prototype response}).
It is worth noticing, that the projection distance $\delta_{\boldsymbol{\Omega}}^{2}\left(\mathbf{x},\mathbf{w}\right)$
from (\ref{eq:reformualted-Omega-diss}) or its nonlinear counter
part $\mu_{\boldsymbol{\Omega}}^{2}\left(\mathbf{x},\mathbf{w}\right)$
from (\ref{eq:nonlinear projection distance}) can also be used in
RSLVQ. Furthermore, the WTA-rule in RSLVQ results into a maximum probability
decision regarding the probability vector $\hat{\mathbf{p}}\left(\mathbf{x}\right)$
as it is applied for FCLs. 

\paragraph{GLVQ}

As mentioned previously, a prominent LVQ network is the generalized
LVQ architecture (GLVQ) with a loss function approximating the overall
classification error \cite{sato96a}, which realizes a hypothesis-margin
maximizer \cite{Crammer2002a}. Given a prototype response vector
$\mathbf{d}\left(\mathbf{x}\right)$ regarding an input $\mathbf{x}$,
the loss is evaluated as 
\begin{equation}
\frac{d_{c\left(\mathbf{x}\right)}\left(\mathbf{x}\right)-d_{\bar{c}\left(\mathbf{x}\right)}\left(\mathbf{x}\right)}{d_{c\left(\mathbf{x}\right)}\left(\mathbf{x}\right)+d_{\bar{c}\left(\mathbf{x}\right)}\left(\mathbf{x}\right)}\label{eq:glvq loss}
\end{equation}
where $\bar{c}\left(\mathbf{x}\right)$ is the class of the closest
prototype of an incorrect class regarding $\mathbf{x}$. This loss
is not comparable with the commonly used NN losses. However, as already
discussed in Sec.$\,$\ref{subsec:Omega distance and FCL}, applying
the projection distance $\delta_{\boldsymbol{\Omega}}^{2}\left(\mathbf{x},\mathbf{w}\right)$
from (\ref{eq:reformualted-Omega-diss}) or its nonlinear counter
part $\mu_{\boldsymbol{\Omega}}^{2}\left(\mathbf{x},\mathbf{w}\right)$
from (\ref{eq:nonlinear projection distance}) in LVQ results in a
network model with two FCLs which can be trained by using (\ref{eq:glvq loss}).

\subsection{VQ layers as prototype convolution operation\label{subsec:VQ-layers-convolutionals}}

Similar to Sec.~\ref{sec:A-different-view-on-CNNs}, we now suppose
a convolutional kernel $\mathbf{k}\in\mathbb{R}^{w_{f}\times h_{f}\times c}$.
However, rather than considering the kernel $\mathbf{k}$ as a learnable
filter, we consider the kernel as a learnable (kernel-)\,prototype.
Then, the filter response of the convolution operation is the distance
between the sliding window and the kernel-prototype. We denote the
convolution operation in combination with a kernel-prototype using
the $\circledast$ sign and refer to it as \emph{kernel-prototype
convolution}.

Using the Euclidean distance, expressed in terms of dot products (\ref{eq:identity-of-dE-to-dot-product}),
the kernel-prototype convolution $\mathbf{x}\circledast\mathbf{k}$
at image position $i,j$ yields
\[
\left.\mathbf{x}\circledast\mathbf{k}\right|_{ij}=\left\Vert \tilde{\mathbf{x}}_{ij}\right\Vert _{2}^{2}-2\tilde{\mathbf{x}}_{ij}^{\textrm{T}}\tilde{\mathbf{k}}+\left\Vert \tilde{\mathbf{k}}\right\Vert _{2}^{2}
\]
and for the whole image
\begin{equation}
\mathbf{x}\circledast\mathbf{k}=\mathbf{x}^{2}\ast\mathbf{1}-2\mathbf{x}\ast\mathbf{k}\oplus\left\Vert \mathbf{k}\right\Vert _{2}^{2}\label{eq:kernel-proto-convolution}
\end{equation}
where $\left(\cdot\right)^{2}$ is the component-wise square, $\mathbf{1}$
is a kernel of ones with the shape equal to $\mathbf{k}$ and $\oplus$
is the component-wise addition. This equation can be efficiently evaluated
and hence, a kernel-prototype of arbitrary kernel-size can be convoluted
over an input. In the previous example, the kernel-prototype is equivalent
to a $w_{f}\cdot h_{f}\cdot c$ dimensional vector. Thus, the kernel-prototype
convolution measures the distance to the kernel-prototype in a $w_{f}\cdot h_{f}\cdot c$
dimensional space. 

For a kernel-prototype convolution the number of filters $N_{f}$
equals the number of kernel-prototypes $N_{W}$. Each feature map
in a kernel-prototype convolution therefore corresponds to the distance
of the input to a different kernel-prototype. To better reflect the
output of the kernel-prototype convolution, we refer to each feature
map as \emph{dissimilarity map} and the resulting whole feature stack
as \emph{dissimilarity stack}. The encoding of a given input into
a dissimilarity representation is equivalent to a counter-propagation
network representation \cite{hecht-nielsen87a,hecht-nielsen88a}.

To strengthen the relation between the normal convolution and the
kernel-prototype convolution a bias term can be added. For this purpose,
we consider the equation of a $n$-ball
\[
B_{r}\left(\mathbf{w}\right)=\left\{ \mathbf{x}\in\mathbb{R}^{n}|d\left(\mathbf{x},\mathbf{w}\right)<r\right\} 
\]
with radius $r$ around the (usual) prototype $\mathbf{w}$. Thus,
the expression 
\[
r^{2}-\mathbf{x}\circledast\mathbf{k}\begin{cases}
>0 & \textrm{if the vector of the sliding window is inside \ensuremath{B_{r}\left(\tilde{\mathbf{k}}\right)}}\\
\le0 & \textrm{else}
\end{cases}
\]
returns a positive value if a sliding window is inside the spanned
$n$-ball of a (usual) prototype and a negative value otherwise. The
evaluation of multiple filters $N_{f}$ can therefore be seen as determining
for each point in which $n$-ball it is located. By only assigning
a positive score when a point is within a $n$-ball, the kernel-prototype
convolution can be interpreted as a quantization. These properties
will be further explored in the following sections. 

It is important to note, that it is still possible to work directly
on the dissimilarity stack. Moreover, the application of a convolutional
filter, with a linear activation before a kernel-prototype convolution
without a bias, is similar to measuring with an Omega dissimilarity,
as discussed in section \ref{subsec:Omega distance and FCL} regarding
two FCLs.

\section{Potential benefits}

PBs and NNs have different benefits over each other in classification
tasks. In this section we will provide an overview of how we think
these benefits will transfer when PBs and NNs are combined. As this
is still an active area of research we do not intent this to be a
complete overview. Where possible, we will provide examples of how
these benefits can be experienced in an experimental setting.

\subsection{Feature extraction pipeline for LVQ \label{subsec:Feature-extraction-pipeline}}

In terms of accuracy, it is not necessary to discuss about which machine
learning method has a broader application. NNs are impressively powerful
in solving real world image tasks like object detection or object
classification. Due to the high complexity of such tasks, plain LVQ
cannot provide a satisfactory accuracy. The power of NNs results from
their usually underlying deep architectures and the training of the
feature extraction and the classification layers in an end-to-end
fashion. The joint training of these two parts is usually not considered
in PBs. 

However, using the methods described above we were able to train a
NN with prototype-based layers in an end-to-end fashion on both MNIST
and Tiny ImageNet. For MNIST a relatively small network was trained
to a validation accuracy of $99\%$. For Tiny ImageNet\footnote{Tiny ImageNet is a downsampled subset of the ImageNet dataset, see
\url{https://tiny-imagenet.herokuapp.com/}} an altered version of the classic ResNet with $50$ layers was adapted
to use a RSLVQ classification layer. We were able to train the $200$-class
dataset achieving a validation accuracy of $67.4\%$. At the same
time we were able to reduce the number of parameters by $25\%$ by
replacing the FCLs in the ResNet by the RSLVQ architecture. This is,
to the best of our knowledge, the first time that a prototype-based
classifier was trained in end-to-end fashion on a dataset with such
a large number of classes.

\subsection{Distance-based interpretations in NNs}

Even though the research community around LVQ and VQ is relatively
small compared to the NN society, it is a very active area of research
and the researchers have produced useful techniques and interpretations
around the methods. By injecting PBs into NNs it is possible to transfer
the interpretation techniques from PBs in NNs. There are several discussions
about the benefits of interpretable NNs. One might be simply the fact
of human nature: our aim to understand the things surrounding us.
From the perspective of function applications in safety critical system
like autonomous driving, we want to certify functions and want to
define validation and verification standards. Of course, this requires
some understanding and interpretability of the model in use. Otherwise
it is hard to define the positive operational range of the function. 

\paragraph{Definition of the distance and dissimilarity measure}

The simplest and most powerful property of PBs is their clear geometrical
motivation and interpretation. A distance is a well defined expression,
capturing a clear definition about what is close and allowing for
a direct ordering of points in terms of closeness. Mathematically
speaking, the lower bound of distances is well-defined, it is simply
zero. In traditional NNs there is not such a strong interpretation
of the layer outputs. Moreover, the inner product is unbounded, which
may lead to an unbounded output for special choices of activation
functions like linear or ReLU. Using the proposed kernel-prototype
convolutional layers instead of traditional layers can bring this
strong definition and bound into the NNs.

For example in NNs it is hard to get an understanding of what the
filter response of a convolutional layer is or what it represents.
Using a kernel-prototype convolution enables to find a stronger interpretation:
The filter response of a kernel-prototype convolution conveys the
distance between the considered convolution window and the prototype.
In addition to this interpretation ability for the response, the suggested
approach allows to compare filter responses with each other. This
makes it possible to detect which features are most appropriate for
a certain pixel position. 

\paragraph{Class-typical prototypes}

In traditional prototype-based learning methods the prototypes are
supposed to capture the ``ideal'' representation of the data points
of a certain class. A prototype can be seen as an abstraction of the
class that it is demanded to represent. It yields a good approximation
of the data contained in the belonging receptive field. Hence, a prototype
can be interpreted in the same way as the original data. An example
for this is speech recognition: if an LVQ model was trained to recognize
certain words, the resulting prototypes are easy to understand because
they represent speech. This is particularly the case if median variants
of PBs, which restrict the prototypes to be data points, are used
\cite{Villmann2015c}. 

However, in an end-to-end approach with trainable feature extraction
and a final LVQ layer, this property can be lost. This is not solely
an issue of LVQ in combination with end-to-end feature extractors
as it can also occur in plain LVQ. It is called the trade-off between
generative and discriminative prototypes \cite{Villmann2014m}. Generative
prototypes are prototypical for the data and lie inside the point
cloud of their class. Discriminative prototypes can lie outside the
class's point cloud but maintain a better decision border than when
they lie inside.

Adding a regularization term to the used cost function can force the
prototype classifier to be class-typical and we can alleviate this
issue \cite{OehlerGray:CombinedImageCompressionandClassification:IEEETransPAMI1995}.
Furthermore, if the data mapping components like (\ref{eq:nonlinear projection distance})
or any deep preprocessing network are included in the LVQ method,
the regularization term can be used to map the data in such a way
that data points within a receptive field follow a Gaussian distribution
in the projection (feature) space. In that case, the Kullback-Leibler
divergence between the projected local data distribution inside a
receptive field of a prototype and a Gaussian in the projection space
should be minimized.

First experimental numerical results show that using this latter regularization
approach forces the prototypes to be both generative and discriminative
in the projection space at the same time.

\paragraph{Feature importance}

GMLVQ is known for a strong interpretability property of the $\boldsymbol{\Omega}$-matrix.
The product matrix $\boldsymbol{\Lambda}=\boldsymbol{\Omega}^{T}\boldsymbol{\Omega}$
in (\ref{eq:omega-diss}) is denoted as classification correlation
matrix \cite{Villmann2014e}. A trace-normalization preserves $\boldsymbol{\Lambda}$
to be a matrix where the resulting entries are understood as correlation
values of the feature space dimensions. The correlation information
is used to apply precise feature selection and to perform a pruning
after the training. This normalization acts as a constraint during
the training. Using the outlined relation of the dissimilarity $\delta_{\boldsymbol{\Omega}}^{2}\left(\mathbf{x},\mathbf{w}\right)$
from (\ref{eq:reformualted-Omega-diss}) in FCLs, more interpretability
can easily be gained in the weight matrices of these FCLs, particularly
if a deep structure is considered to feed the FCLs. 

\paragraph{Interpretation techniques for NNs}

Recent results have turned NNs from a black-box into a gray-box, relaxing
the accusation that NNs are not interpretable \cite{olah2018the}.
A wide range of powerful tools to analyze convolutional layers and
highlight important regions in an image have become available. With
small modifications the same techniques can also be available for
PBs. This has already be shown in \cite{Villmann2018i}, where decoder
networks were successfully applied to decode a vectorial output (e.\,g.
the prototypes) into a representation in the input space. 

\subsection{Prototype-based robustness in NNs\label{subsec:Prototype-based-robustness}}

The search for networks that are robust against adversarial attacks
has grown a lot in recent years. Nevertheless, almost each defense
can still be be broken with a suitable attack\cite{AthalyeEtAl:ObfuscatedGradientsFAlseSenseSecurity:ICML2018}.
This observation seems to be plausible under the scope of the No-Free-Lunch
theorem \cite{WolpertNoFreeLunch1996}. As mentioned earlier however,
PBs are known for being robust. By using prototype-based layers in
multilayer networks we therefore expect the network to become more
robust against adversarial examples. 

This expectation is mathematically supported by strong theoretical
results regarding the convergence of prototype-based methods \cite{BiehlDynamicsofLVQ_JMLR2007,WitoelarESANN2009},
hypothesis margin optimization by LVQ networks \cite{Crammer2002a}
and other aspects like information theoretic properties, Bayesian
decision theory for LVQ networks and more \cite{Villmann2006u,PrincipeSpringerbook2010,Torkkola2003a,VailayaRejectOptionsBayesianVQClassification2000}.
These results do not improve the robustness of the LVQ methods automatically,
but the theoretical results provide mathematical certainty.

Below we will discuss one property of PBs and one commonly used approach
that we think can contribute to the robustness.

\paragraph{Homogeneity}

As outlined previously, LVQ networks provide a Voronoï tessellation
of the input/\,feature space according to Eq.~(\ref{eq:Voronoi cell})
in dependence on the prototype distribution. Inside each Voronoï cell
$V_{k}$ the class decision is homogeneous according to the class
assignment rule (\ref{eq:LVQ class mapping}). The local $n$-balls
$U_{\varepsilon\left(k\right)}\left(\mathbf{w}_{k}\right)$ around
a prototype as approximations of Voronoï cell $V_{k}$ are defined
in such a way that
\[
\varepsilon\left(k\right)=\textrm{argsupre mum}_{\varepsilon}\left\{ U_{\varepsilon\left(k\right)}\left(\mathbf{w}_{k}\right)\subset V_{k}\right\} 
\]
is valid, meaning that the maximum $n$-ball is fully contained in
the Voronoï cell $V_{k}$. Then the $n$-balls are disjunctive, i.e.
$U_{\varepsilon\left(k\right)}\left(\mathbf{w}_{k}\right)\cap U_{\varepsilon\left(j\right)}\left(\mathbf{w}_{j}\right)=\emptyset,\forall k\neq j$
and the classification decision inside each ball is homogeneous. 

We believe and argue that the verification and validation of safety
critical functions based on classifier decisions will be much more
clear (and therefore much more accepted), if a mathematically precise
description of regions with homogeneous classification decisions is
possible. This argumentation is underpinned by several recently published
proofs for robustness of NNs, where approximations of the $n$-balls
are used with the underlying assumption regarding the valid classification
decision homogeneity \cite{HeinAndriushchenko:FormalGuaranteesOfClassifierRobustnessAgainstAdversarialAttacks:NIPS2017,croce2018provable}. 

\paragraph{WTA and hard assignments}

The WTA-rule commonly used in PBs combined with the crisp classification
assignments offers a level of mathematical abstraction that is not
found comparably in NNs. We believe that without a certain level of
abstraction allowed by homogeneous regions, small perturbations in
the input will result in a large difference in the output. 

For this reason we experimented with hard assignment activations regarding
the kernel-prototype convolutions in our recent simulations: the respective
outputs are binary maps of zeros and ones, so far resulting in promising
results. Additionally, the prototype-based classification layer also
adds a certain level of abstraction. The ResNet with $50$ layers
discussed in Sec.~\ref{subsec:Feature-extraction-pipeline} was ranked
top 20 of overall $312$ participants in the NeurIPS'2018 adversarial
vision challenge, in which robust models were validated using novel
adversarial attacks. 

\subsection{Reject options in NNs }

A further issue regarding robustness in NNs is the rejection of unclassifiable
data, such as adversarial examples that can no longer be recognized.
While a lot of work has been done on outlier detection in NNs \cite{UrahamaFurukawa:GradienDescentNPCwithReject:PattRec1995,TaxDuinGrowingMultiClassClassifierwithRejectPRL2008,LandgrebeDuin:ClasssificationRejectOption_PRL2006,HerbeiWegkampClassifcationwithRejectCJourStat2006,YuanWegkampClassificationRejectConvexRiskMinimizationJMLR2010,BartlettWegkampClassificationRejectOptionHingeLossJMLR2008,deStefanoRejectOrNotRejectforNeuralClassifersIEEE2000},
rejecting those outliers is not common in the defense against adversarial
examples. On the other side outlier rejection is a very common topic
in prototype-based learning \cite{FischerHammerWersing:EfficientRejectionStrategies:Neurocomputing2015,HellmanNNClassifierwithReject1970}.
Most of these reject strategies are based on the the fundamental property
of dissimilarity measures to always be greater or equal zero and that
$\mathbf{x}=\mathbf{y}$ implies $d\left(\mathbf{x},\mathbf{y}\right)=0$.\footnote{Other properties of distance measures can be relaxed for general dissimilarity
measures \cite{Villmann2017b}.} In addition to this there are also some reject strategies used in
prototype-based learning based on Bayesian theory and proven to be
Bayes optimal \cite{TradeoffAccuracyRejectChow1970,VailayaRejectOptionsBayesianVQClassification2000}. 

We have so far experimented with applying two reject strategies from
LVQ in NNs using prototype-based layers. The first one spans $n$-balls
around the prototypes and considers points outside the balls as outliers,
see \cite{FischerHammerWersing:EfficientRejectionStrategies:Neurocomputing2015}
for a possible termination of the radii. The second one uses a misclassification
loss to train the network over minimizing the classification error
rate and the reject rate \cite{Villmann2016a}. The idea is to introduce
misclassification costs $\lambda_{e}$, reject-costs $\lambda_{r}$
and a reject strategy in the network. Via tuning of the network parameters,
the network learns to optimize a trade-off between misclassifying
and rejecting inputs. 

\subsection{Shared losses and regularizations between PBs and NNs}

The NN community is open to new developments, even without a precise
mathematical foundation being presented if a benefit in performance
can really be expected. Entrusting empirical justifications, the community
allows quick iterations and improvements to be made, which is, in
our view, an important factor in the recent success of NNs. It allowed
researchers to think about specialized loss functions, regularization
terms or particular optimizers for the considered problem, which can
be easily adapted to new problems. 

In contrast, most LVQ variants are optimized according to the underlying
GLVQ-loss, which is a smooth approximation of the classification accuracy
but, as already mentioned above, corresponds to a hypothesis margin
maximizer. Probabilistic variants like RSLVQ rely on log-likelihood
ratios, which turn out to be equivalent to the cross entropy loss
under mild assumptions \cite{Villmann2018b}. We wish that the PB
community is inspired by the rapidly developing NN-world to think
towards adaptation for vector quantization models. A respective example
is the incorporation of dropout techniques for GMLVQ for stability
analysis \cite{Villmann2018f}. 

\paragraph*{Nonlinearities}

Nonlinearities in the form of activation functions play a key role
in both PBs and NNs. Both methods however rely primarily on different
activation functions. For NNs the ReLU function is mostly used for
intermediate layers and the softmax function for the final classification
layer. For PBs, the focus has been primarily on the sigmoid and linear
activation function. However, recently the swish function has been
showed to slightly improve the classification of many standard datasets
over the ReLU function in NNs \cite{RamachandranEtAl:SearchingForActivationFunctions:arXiv2018}.
This shows the benefit that can be gained from switching to a different
activation function and is a motivation to incorporate activation
functions such as ReLU in PBs and vice versa. Additionally, the use
of ReLU for GLVQ classification layers in NN has already been shown
\cite{YangEtAl:ClassificationWithConvolutioanlPrototypeLearning:CVPR2018}.

Instead of using the NN inspired activation functions for the final
classification only, they can also be applied to the kernel-prototype
convolutional layer. An example of this is the usage of a ReLU function
to clip the convolutional layer with bias. By doing so, positive values
are assigned to points inside the $n$-ball $B_{r}\left(\mathbf{w}_{k}\right)$
of a prototype and zeros to points outside. The largest value is then
defined by the bias, with larger values when the vector is closer
to the prototype. Furthermore, we experiment with applying the softmax
activation function over the channels of the dissimilarity stack to
assign cluster probabilities to each pixel position. Equivalently,
we use the sigmoid activation to get cluster possibilities instead
of probabilities. This way, a value greater than $0.5$ is assigned
if a point is inside the $n$-ball of a prototype and lower otherwise.

\paragraph{Regularization}

Similar to nonlinearities, using prototype-based layers allows to
share regularization techniques developed for NNs in PBs, like the
$l_{1}$-norm regularization term known from NNs to the bias term
of kernel-prototype convolutions. The idea is to keep the trust region
around a prototype as small as possible. During the training the network
automatically turns prototypes off which are not used in the following
layers. After the training the prototypes can be removed without any
effect on the network performance. Similar approaches were presented
in \cite{Villmann2018e} for sparse feature utilization in GMLVQ. 

The other way around, regularization techniques known in GMLVQ can
provide new aspects for NNs. An example of this is to apply the eigenvalue
regularization known for GMLVQ (see \cite{Villmann2013e0,Villmann2010l})
for a two-layer network of FCLs, which can potentially reduce the
over-simplification effect in the projection space.

\subsection{Arbitrary input dimensions and structured data for NNs and vector
quantization}

In \cite{HintonEtAl:DynamicRoutingCapsules:NIPS2017} a capsule concept
was proposed to work on tensor signals between neurons/\,perceptrons.
The output over all neurons of a FCL can be considered as a vector.
In a capsule network, the entities that are transmitted along a connection
between two capsules are arbitrary tensors and thus, the collected
output of a capsule layer is a tensor and therefore not longer restricted
to be a vector. This concept allows to think about new processing
steps inside and between capsules and hence to go beyond vectorial
operations. This is however not a new concept in machine learning,
but rather a well established property of PBs: a simple example is
to use a matrix as the input to a VQ network, then the prototypes
are matrices as well and hence, a respective matrix dissimilarity
has to be chosen \cite{Villmann2016f}. However, using more complex
dissimilarity measures it is also possible to model prototypes as
affine subspaces of the input space \cite{Villmann2016h,Villmann2016j}.

\section{Some tricks to successfully train prototype-based neural
network layers\label{sec:Some-tricks-to}}

Merging both methods does not only bring their benefits but also their
difficulties, e.\,g. that the PBs are hard to train compared to NNs.
However, a variety of techniques have already been developed in both
research areas that can also be used in the merged methods to alleviate
(part) of these difficulties.

\paragraph*{Prototype initialization}

The training of PBs depends usually on a good initialization \cite{nishida01a,QinSuganthanInsensitiveLVQPatternRecognition2004,xiong04a_bibuniq_3366}.
For this reason, instead of random initialization, we use two different
strategies which are frequently used to train VQ/\,LVQ networks:
\begin{itemize}
\item the prototypes are initialized as randomly chosen input vectors
\item the prototypes are initialized as centers of a k-means algorithm or
neural gas 
\end{itemize}
A similar strategy works for a proper initialization of the biases
in a prototype convolution. 

\paragraph*{Prototype regularization}

As noted in \cite{martinetz93d}, one problem in learning prototypes
is that they often remain unchanged because of an initialization in
data regions of low density. To avoid this, neighborhood cooperativeness
has been established using an external grid structure following the
dynamics of learning in neural maps in cortical brain areas \cite{kohonen95a}.
A similar concept without external grid is the neural gas quantizer
proposed by \noun{Martinetz} and \noun{Schulten} with a winning rank
dependent neighborhood cooperativeness \cite{martinetz93d}. We use
this method in our experiments by defining the gradient back-flow
of VQ layers in the update step as rank dependent. 

For some of the regularization and loss terms described in this report
it is needed to estimate the data distribution over the training dataset.
Since the network is optimized by stochastic gradient descent learning,
the statistics over the whole dataset cannot be estimated during run-time.
This can be compensated via moving averages/\,moving variances. Additionally,
we frequently prefer to perform a zero de-biasing according to \cite{KingmaBa:AdamMethodStochasticGradient:ICLR2015}
to avoid biased gradients.

\paragraph{Hard assignment derivative}

As discussed in Sec.~\ref{subsec:Prototype-based-robustness}, we
carried out experiments using discrete outputs for the prototype-based
layers. Unfortunately, a network with discrete output values is not
differentiable. Thus, we needed to define approximated gradients.
The hard assignments are defined and trained in the following way:
\begin{enumerate}
\item In case of a prototype convolution without bias, the output value
of each pixel is defined as one for the closest prototype and zero
for all the other prototypes. Thus, the dissimilarity stack becomes
a stack of unit vectors. This hard probability assignment can be approximated
by applying the function (\ref{eq:neg_softmax}) with a kernel width
$\sigma$ equivalent to the sigmoid function. If $\sigma$ tends to
be zero, the $\textrm{\textrm{softmax}}$ function approximates a
hard probability assignment.\label{enu:proto-con-softmax}
\item In case of a prototype convolution with bias, the Heaviside step function
is applied as activation for each dissimilarity value. Thus, points
inside the $n$-ball $B_{r}\left(\mathbf{w}_{k}\right)$ of a prototype
are assigned to one and zero otherwise. Again, this hard possibility
assignment can be approximated by applying the sigmoid function.\label{enu:proto-conv-sigmoid}
\end{enumerate}
In the forward pass of the network, the hard output assignments are
transmitted to the next layer. In the backward pass of the network,
the gradient is approximated by the $\textrm{\textrm{softmax}}$ and
sigmoid function respectively. By adaption of the kernel width $\sigma$
the quality of these approximations can be controlled. This strategy
is similar to the gradient-straight-through approach applied in \cite{OordEtAl:NeuralDiscreteRepresentationLearning:NIPS2017}.
Theoretically, the approach \ref{enu:proto-conv-sigmoid}. is capable
to handle more complex tasks. The reason is that in this approach,
the encoding for the following layer is not restricted to unit vectors.
More precisely, in approach \ref{enu:proto-con-softmax}. $N_{W}$
different encodings can be created in the probability stack whereas
the number of distinguishable encodings in \ref{enu:proto-conv-sigmoid}.
is $2^{N_{W}}$.

\section{Related work}

One of the first contributions reporting an approach for the fusion
of LVQ with NNs is \cite{deVries:DeepLVQ:ESANN2016}. Later on, the
idea was formulated more precisely \cite{Villmann2017h}. In \cite{YangEtAl:ClassificationWithConvolutioanlPrototypeLearning:CVPR2018},
a fused network was applied to train a network on MNIST and Cifar10.
They showed a way how a regularization term can be used to get a generative
and discriminative model at the same time. Moreover, they exemplary
applied a reject strategy and showed that the concept of incremental
learning is also applicable. 

The definition of a network directly on the discrete cluster representation
of a VQ layer was first mentioned in the VQ-Variational-Autoencoder
(VQ-VAE)  \cite{OordEtAl:NeuralDiscreteRepresentationLearning:NIPS2017}.
There, the output of the VQ is a map where a pixel has the integer
number $k$ of the closest prototype. Moreover, the output is defined
as the latent space of the VQ-VAE. The method was trained via the
gradient-straight-through method. The obtained results are promising.
Nevertheless, the authors were unable to train the model from scratch
even with soft-to-hard assignment proposed in \cite{AgustssonEtAl:SoftHardVQForEnd2EndLearning:NIPS2017}.
We made similar observations like in \cite{OordEtAl:NeuralDiscreteRepresentationLearning:NIPS2017}
for our simulations performed so far. If we train using soft-to-hard
assignments, the model is able to invert the continuous relaxations.
We also observed that training by a simple gradient-straight-through
approximation is not stable, when the model is trained from scratch.
However, our proposed gradient approximation is working well for many
performed settings, either to train from scratch or on a pre-trained
network.

In the VQ-VAE, the closest prototype at each pixel position is used
directly as the input to the decoder. This architecture can easily
be fitted into our framework: it is equivalent to applying a $1\times1$
kernel-prototype convolution without bias to the final layer of the
encoder, followed by a hard class assignment (see Sec.~\ref{sec:Some-tricks-to},
method \ref{enu:proto-con-softmax}). Afterwards a $1\times1$ transposed
convolution has to be applied over the unit vector stack with the
number of filters equaling the prototype dimension. If the prototypes
are used as inputs for the decoder, the weights in the $1\times1$
convolution can be defined as the prototypes from the kernel-prototype
convolution. Thus, the two convolutional layers share the weights.
Instead of sharing the weights, we propose to just initialize the
weights of both layers equivalently and then train the kernels independently.
If the prototype representations are appropriate to start the decoding,
we can assume that these weights will be discovered by the training
anyways. In fact, this framework is more generic than the sharing
weights approach, because it does not raise the question how the prototypes
are pushed to the decoding network if the kernel-prototype convolution
is defined via a kernel size greater than $1\times1$ to include neighborhood
relations in the quantization process.

In the paper \cite{Villmann2018i} a framework is presented to fuse
VQ/\,LVQ with capsule networks \cite{HintonEtAl:DynamicRoutingCapsules:NIPS2017}.
We spend a lot effort to analyze this network architecture and corresponding
routing mechanism. First of all, we found that this network is not
able to transmit the message that an attribute is not present in the
given representation to the next layer. For example, assume the problem
of discrimination between cats and boats. Obviously, the presence
of the attribute water is of interest for the class boats and its
absence is important for the class cats. Since the routing is driven
by preferring signals with a small distance, the signal from the water
prototype indicating that water is not present (big distance) will
be vanished in the routing process by assigning small routing probabilities
to the next layer. Thus, the network decision is driven by present
entities only. Moreover, we observed that the network learns to ignore
the routing process by pushing the whole signal flow over just one
prototype. The following layers were just working on this one hot
signal (or a few of them) instead of using the whole network capacity. 

\section{Conclusion}

In this conceptual paper we presented relations between VQ/\,LVQ
and NN approaches for classification and how they fit into each other.
We summarized important concepts from both theoretical fields, which
are potentially not common/unknown in the other field. As major contribution
we showed, how an Euclidean distance calculation can be expressed
in terms of convolution operations in the NN sense. This and the computation
of the Euclidean distance in terms of a dot product are essential
steps towards efficient computation schemes for prototype-based neural
network layers. Using these approaches we were able to train a ResNet50\,+\,LVQ
network on Tiny ImageNet with 200 classes without any scalability
problems. 

Right now, we are evaluating architectures with our proposed layers
and training pipelines. At several steps in the paper we mentioned
current results without presenting precise quantities because the
evaluation is still going on. The presentation of the experimental
results will be part of our upcoming contribution.

Overall, we hope that this overview is stimulating for researchers
from both communities to study together on fused models and to discover
how established frameworks from one method can be incorporated into
the other. We also hope that this contribution serves a good starting
point for research and cites the recent trends from both sides. 

\bibliographystyle{unsrt}

\end{document}